\documentclass{article}

\usepackage{arxiv}

\usepackage{amsmath,amssymb,amsthm}
\usepackage{mathtools}
\usepackage{xspace}
\usepackage{epsfig}
\usepackage[vlined,linesnumbered,ruled,titlenotnumbered]
{algorithm2e}
\usepackage{color}
\usepackage{url}
\usepackage{graphicx,dsfont}
\usepackage{tikz}
\usepackage{float}
\usepackage{longtable}
\usepackage{booktabs}
\usepackage{makecell}
\usepackage{multicol}
\usepackage{multirow}
\usepackage{colortbl}
\usepackage{tikz}
\usepackage{csquotes}

\newcommand{\ignore}[1]{}
\newcommand{\wdtsps}{W-TSP}
\newcommand{\wtour}{\ensuremath{\mathcal C}}
\newcommand{\wcity}{\ensuremath{\omega}}
\newcommand{\jakob}[1]{\textbf{\textcolor{violet}{Jakob: #1}}}

\begin{document}


\author{
   Jakob Bossek \\
   Department of Computer Science \\
   RWTH Aachen University \\
   Aachen, Germany \\
   \texttt{bossek@aim.rwth-aachen.de}
   \And
   Aneta Neumann \\
   Optimisation and Logistics\\
   The University of Adelaide\\
   Adelaide, Australia \\
   \texttt{aneta.neumann@adelaide.edu.au}
   \And
   Frank Neumann \\
   Optimisation and Logistics\\
   The University of Adelaide\\
   Adelaide, Australia \\
   \texttt{frank.neumann@adelaide.edu.au}
}

\title{On the Impact of Operators and Populations within Evolutionary Algorithms for the Dynamic Weighted Traveling Salesperson Problem}

\maketitle

\begin{abstract}
Evolutionary algorithms have been shown to obtain good solutions for complex optimization problems in static and dynamic environments. It is important to understand the behaviour of evolutionary algorithms for complex optimization problems that also involve dynamic and/or stochastic components in a systematic way in order to further increase their applicability to real-world problems.
We investigate the node weighted traveling salesperson problem (W-TSP),
which provides an abstraction of a wide range of weighted TSP problems, in dynamic settings.
In the dynamic setting of the problem, items that have to be collected as part of a TSP tour change over time. We first present a dynamic setup for the dynamic W-TSP parameterized by different types of changes that are applied to the set of items to be collected when traversing the tour. Our first experimental investigations study the impact of such changes on resulting optimized tours in order to provide structural insights of optimization solutions.
Afterwards, we investigate simple mutation-based evolutionary algorithms and study the impact of the mutation operators and the use of populations with dealing with the dynamic changes to the node weights of the problem.
\end{abstract}

\section{Introduction}
\label{sec:introduction}

Evolutionary algorithms~\cite{EibenS2015} have been successfully applied to a wide range of complex optimization problems. 
Such optimization problems occurring in real-world applications are often challenging to solve due to a combination of aspects that makes the optimization part challenging~\cite{DBLP:series/sci/BonyadiM16}. On the one hand, real-world optimization problems often consist of a combination of $\mathcal{NP}$-hard combinatorial optimization problems~\cite{DBLP:books/sp/19/BonyadiM0019}. Furthermore, the interactions of these problems imply that the different problems can not be solved in isolation and the most challenging part consists of dealing with the problem interactions~\cite{DBLP:journals/ori/StolkMMM13}.
On the other hand, real-world problems often involve dynamic and/or stochastic components. Dealing with dynamically changing circumstances is essential in order to quickly react to changes in terms of the variability of resources or changes in terms of production costs.

Evolutionary algorithms are able to adapt their solutions to changing circumstances and are therefore well suited in dynamic environments~\cite{DBLP:journals/swevo/NguyenYB12}.
Understanding the behaviour of evolutionary algorithms for complex optimization problems with interacting components that also involve dynamic and/or stochastic components is highly challenging, but also highly important to further increase the applicability of evolutionary algorithms to real-world problems. The aim of this article is to contribute to this direction of research. We investigate the node weight dependent traveling salesperson problem as an abstraction of a wide range of node weighted TSP variants in dynamic settings in order to provide new insights into the optimization behaviour of evolutionary algorithms. Our aim is to understand how important mutation operators commonly used for permutation problems and population impact the ability of evolutionary algorithms to deal with dynamic changes. Understanding these components in a sound manner and being able to point out the impact of different operators and population sizes is crucial before studying more complex algorithms involving crossover and other more sophisticated techniques for dealing with dynamic changes.

\subsection{Related Work}

Our goal is to study dynamic settings and how evolutionary algorithms can cope with them in a systematic way. In order to do this, we restrict ourselves to simple variants of evolutionary algorithms that allow us to study systematically the impact of important operators and components such as mutation operators and populations. It should be noted that there are a variety of studies in the area of dynamic continuous optimisation where researchers developed different types of high performing evolutionary computing approaches~\cite{DBLP:journals/eor/MukherjeeDD16,DBLP:journals/isci/MukherjeePKD14}.

There are different types of TSP variants where the travel cost from $i$ to $j$ depends on some weighting of the given distances.
The traveling thief problem~\cite{DBLP:conf/cec/BonyadiMB13} is such a problem and has attracted significant interest in the evolutionary computation literature in recent years~\cite{Faulkner2015,ElYafrani:2016:PVS:2908812.2908847,ElYafrani2018231,DBLP:conf/gecco/WuP0N18,DBLP:conf/seal/Wu0PN17,DBLP:conf/gecco/WuPN16,Wagner2017ttpalgssel}. It combines two classical $\mathcal{NP}$-hard combinatorial optimization problems, namely the traveling salesperson problem (TSP) and the knapsack problem (KP), into a problem where a TSP tour and a KP packing has to be found such that a goal function combining KP profits and KP weight dependent tour costs is optimized. These interactions of the TSP and KP make the TTP significantly harder to solve than the silo subproblems and important research has been carried out to understand such interactions of weights and distances~\cite{DBLP:journals/swevo/NguyenYB12,DBLP:journals/soco/MeiLY16,WTSPapprox}.

In the TTP, the cost of traveling from city $i$ to city $j$ increases linearly with the decrease of the velocity of the traveling thief. The reduction in velocity on the other hand depends linearly on the increase in weight of the items already collected. Overall, this makes the impact of the weights hard to understand and it seems to be natural to investigate an increase in travel cost that depends linearly on the weight before studying the velocity dependent cost increase in a systematic way. The node weighted traveling salesperson problem (\wdtsps{}) introduced in~\cite{WTSPapprox} provides such a simplification and it has been shown recently that a wide subclass of instances can be approximated within a factor of $3.59$ by adapting approximation algorithms for the minimum latency problem~\cite{latency_best_apx}.

On the other hand, dynamic problems have received increasing attention in the evolutionary computation literature from a theoretical as well as practical perspective in recent years~\cite{DBLP:journals/corr/abs-1806-08547,DBLP:conf/cec/Ameca-AlducinHB18,DBLP:conf/evoW/Ameca-AlducinHN18,DBLP:journals/tec/ChenLY18,DBLP:conf/gecco/YazdaniBO0Y18}. Recently, different types of subset selection problems such as the classical knapsack problem and the optimization of submodular functions with dynamically changing constraints have been investigated~\cite{DBLP:journals/corr/abs-1811-07806,
DBLP:journals/corr/abs-2004-12574}.
Furthermore, stochastic constraints have been investigated for the classical knapsack problem in the static and dynamic settings~\cite{DBLP:journals/corr/abs-2002-06766,DBLP:journals/corr/abs-2004-03205} and investigations have been carried out for the optimization of submodular functions with stochastic constraints~\cite{DBLP:journals/corr/abs-1911-11451,AnetaPPSN2020}.
Understanding the impact of dynamic changes to components of a given problem as well as the ability of evolutionary algorithms to deal with them is currently a very active area of research~\cite{DBLP:journals/tec/YazdaniOBNY20,DBLP:journals/corr/abs-1907-13529}.

\subsection{Our contribution}

In this article, we investigate a dynamic version of the node weighted traveling salesperson problem (\wdtsps{}) introduced in~\cite{WTSPapprox}.
In the \wdtsps{} the travel cost increases linearly with the collected weight and it takes the motivation from TTP as well as other important TSP variants such as the time-dependent traveling salesman problem which involves a weightening of the travel cost in dependence on the preceding part of the salesperson tour~\cite{tdtsp_origin,wrong_tdtsp,tdtsp_genetic}.

In the \wdtsps{}, each node has a given weight that increases the cost of the remaining salesperson tour. In order to establish a dynamic setup for the problem, we model the weight on a node based on a set of items that need to be collected.
In our setting, the set of items that need to be picked up is dynamically changing which implies that a packing plan is given and an optimal tour for such a plan has to be found. The packing plan changes over time which requires that the algorithm has to adapt and reoptimize the tour based on the changes of the packing. Changing the packing plan in the (node) weighted TSP is equivalent to changing the weights on the nodes of an instance and we study the effect of such changes on the problem together with the performance of evolutionary algorithms that reoptimize the given objective function.

We set up a framework for benchmarks where the availability of items varies over time. The framework establishes a general setting for dynamic problems where the availability of items changes over time and we hope this is of general interest in the context of dynamic optimization. Our framework changes the availability of items based on a random process where the initial set of items can be determined as well as upper and lower bounds on the number of items that are available at each time step. Each dynamic change makes on average $c$ percent new items available and another $c$ percent items unavailable while respecting the upper and lower bounds on the total number of available items. Here $c$ is a parameter that determines the magnitude of change. Note that the process induces a fair random walk on the number of items when ignoring the boundary cases. However, the total composition of items that are available can change significantly during this process. We investigate our algorithms for various settings of $c$ and different values for the upper and lower bounds. The other important parameter for our dynamic benchmark setting is the frequency of changes determined by a parameter $\tau$ which determines how many fitness evaluations may be executed until the next dynamic change happens.

We investigate how our dynamic settings change the \wdtsps{} instance and the tour optimized by evolutionary algorithms. Our investigations show how the location of the cities in the resulting tour changes dependent on the node weight changes that have occurred. Afterwards, we investigate the classical $(1+1)$ and a $(\mu+1)$-EA and show when a population is usually helpful with the dynamic changes. Our results show that populations are not helpful if the frequency of changes is too high, i.e. $\tau$ is small. On the other hand, a large value of $\tau$ enables the $(\mu+1)$-EA to make use of its diverse solution set such that the algorithm can deal and cater for changes applied to the problem.

The outline of this article is as follows. In Section~\ref{sec2}, we introduce the dynamic \wdtsps{}. We describe the baseline algorithms used in our experimental investigations in Section~\ref{sec:sec3} and introduce the framework for the dynamic changes in Section~\ref{sec:sec4}. In Section~\ref{sec:sec5}, we carry out investigations that show how the dynamic changes impact the problem instances and resulting optimized solutions. We report on our experimental investigations for baseline evolutionary algorithms in Section~\ref{sec:sec6} and finish with some concluding remarks.

\section{The Dynamic Node-Weighted Traveling Salesperson Problem}
\label{sec2}
We consider the node weight dependent traveling salesperson problem (\wdtsps) introduced in \cite{WTSPapprox}. The input is given as a set of $n$ cities $V =\{1, \ldots, n\}$ together with a weight function $w \colon V \rightarrow \mathds{R}^+$ and a distance function $d \colon V \times V \rightarrow \mathds{R}^+$. For a permutation $\pi$ of $V$,
we define $\wcity_i(\pi)=\sum_{j=1}^i w(\pi_j)$ as the sum of the weights of the first $i$ cities in $\pi$. In \wdtsps{} the goal is to find a permutation $\pi$ of $V$ that minimizes the node weighted traveling cost given as
\begin{align}
\label{eq:wtsp_fitness_function}
\wtour(\pi) = d(\pi_n, \pi_{1})\wcity_n(\pi)+ \sum_{i=1}^{n-1} d(\pi_i, \pi_{i+1})\wcity_i(\pi).
\end{align}
In $\wtour{}$ the distance $d(\pi_i, \pi_{i+1})$ is multiplied by the sum of the weights along the tour until $\pi_i$ is visited. 
Note, that the classical (unweighted) TSP is the special case where $w(\pi_1)=1$ and $w(\pi_i)=0$ for $2 \leq i \leq n$. We always require that each permutation starts with city $1$ and therefore require $\pi_1=1$ for any feasible permutation $\pi$.
 
We consider the following setting of \wdtsps{}. Each city $i$ contains a set $E_i$ of items $e_{ij}$, $1 \leq j \leq m_i$, where $m_i=|E_i|$, and $E_i \cap E_j= \emptyset$ for $i \not = j$. The overall set of items is denoted by $E = \cup_{i=1}^n E_i$ and we denote by $m = |E|$ the total number of items of a given instance. For a given item $e_{ij} \in E$, we denote by $w(e_{ij})$ its weight. In our dynamic setting, items can be active or inactive and we encode this by a bitstring $x \in \{0,1\}^m$ where $x_{ij}=1$ iff item $e_{ij}$ is active. Note, that we are using double subscript notation $x_{ij}$ here to denote the bit for the $j$-th item of city $i$. 
In the case, where we do not refer to items explicitly, we use the notation $x \in \{0,1\}^m$ without explicitly mentioning the locations. This is especially the case when we are referring to the dynamic changes carried out to the set of available items.

Active items have to be collected when traversing the tour and the weight of node $i$ is given as
$$
w(i) = \sum_{j=1}^{m_i} w(e_{ij}) x_{ij}.
$$

\begin{figure*}[ht]
    \centering
    \includegraphics[width=\textwidth, trim=10pt 0.65cm 3pt 0pt, clip]{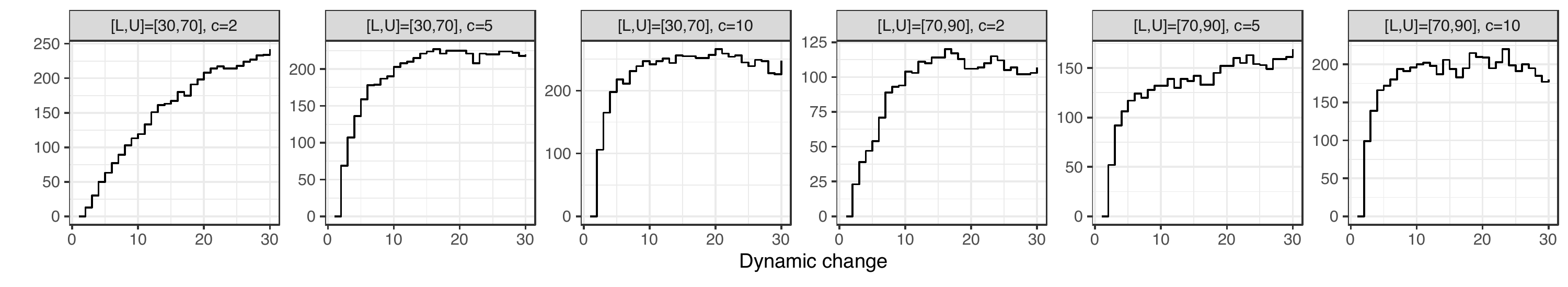}
    \includegraphics[width=\textwidth, trim=10pt 0.65cm 3pt 0pt, clip]{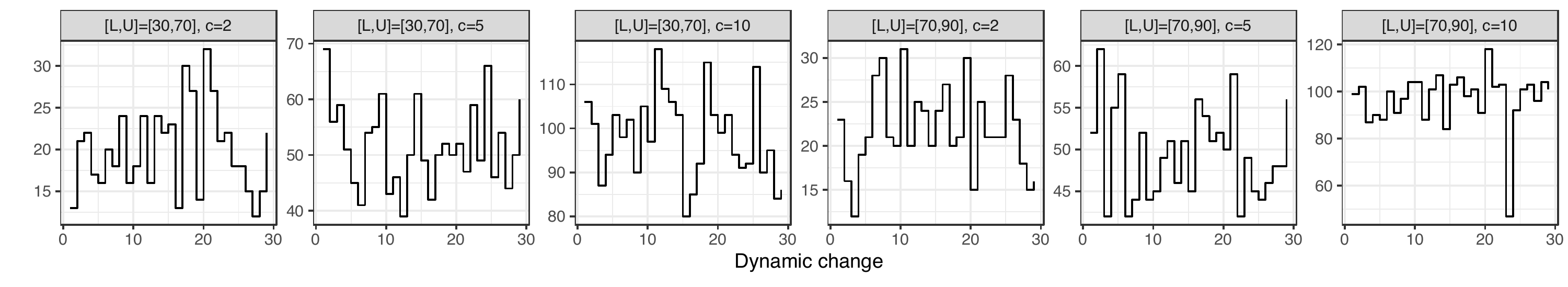}
    \includegraphics[width=\textwidth, trim=10pt 7pt 3pt 0pt, clip]{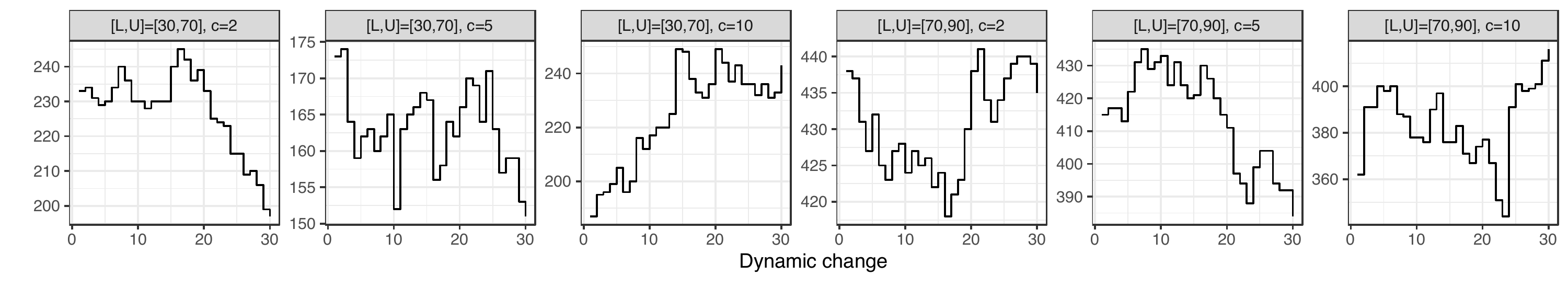}
    \caption{Visualization of the Hamming-Distance $HD(x^0, x^t)$ to the initial packing $x^0$ (top row), the Hamming-distance $HD(x^{i}, x^{i+1})$ of subsequent packings (middle row) and the number of active items in the course of consecutive epochs $i = 0, \ldots, 30$ (bottom row) for an instance with $500$ items. Data is split by the dynamism parameters $[L, U]$ and $c$.}
    \label{fig:dynamic_setup}
\end{figure*}
 
In our dynamic setting the bitstring $x$ changes over time which implies that items can change their status from active to inactive or vice versa. This directly changes the weight of the nodes where such changes occur. 

Let $x^t \in \{0,1\}^m$ be the bitstring determining the set of available items at time step $t$. The weight associated with node $i$ is then given as
$$
w^t(i) = \sum_{j=1}^{m_i} w(e_{ij}) x^t_{ij}
$$
and the cost of a tour $\pi$ is given as
\[
\wtour^t(\pi) = d(\pi_n, \pi_{1})\wcity^t_n(\pi)+ \sum_{i=1}^{n-1} d(\pi_i, \pi_{i+1})\wcity^t_i(\pi)
\]
where
$\wcity^t_i(\pi)=\sum_{j=1}^i w^t(\pi_j)$.

\section{Considered Algorithms}
\label{sec:sec3}
We consider simple evolutionary algorithms that work on permutations of the given $n$ cities. The goal is to minimize the fitness function \wtour{} (see Eq.~(\ref{eq:wtsp_fitness_function})) and dynamic changes are applied to the items that are collected when traversing the tour.

The ($\mu+1$)-EA (see Algorithm~\ref{alg:ea}) starts with a population $P$ consisting of $\mu$ permutations and selects in each iteration one permutation $\pi \in P$ for mutation. The offspring $\pi'$ of $\pi$ replaces $\pi$ if its weighted tour length is not larger than the one of $\pi$. The process is iterated until a stopping criterion is fulfilled.  
We also consider the special case $\mu=1$ which results in the classical $(1+1)$-EA often considered as a baseline algorithm in theoretical studies on the runtime analysis of evolutionary algorithms.
\begin{algorithm}[t]
    Choose a multi-set $P$ of $\mu$ random permutations on $V$.\\
    Choose $\pi \in P$ uniformly at random.\\
Produce $\pi'$ from $\pi$ by mutation.\\
    If $\wtour(\pi') \leq \wtour(\pi)$, set $\pi:=\pi'$.\\
    If termination condition is not satisfied, go to 2. 
  \caption{$(\mu+1)$-EA.}
  \label{alg:ea}
\end{algorithm}
As mutation operators, we consider classical mutation operators for permutation representation: inversion, exchange, and jump which are frequently used for permutation problems (see, e.g. \cite{EibenS2015}). All three operators select two positions $i,j \in \{1, \ldots, n\}$ in a given permutation $\pi$. Inversion mutation reverts the sequence in-between positions $i$ and $j$; this is actually a so-called 2-OPT move which, in its general extension $k$-OPT with $k\geq2$ is a basic ingredient of many TSP-heuristics~\cite{helsgaun_general_2009}. Exchange-mutation simply exchanges the elements at the selected positions while jump-mutation moves the element in position $i$ just before/after position $j$; all other elements between $i$ and $j$ are shifted.
Our goal is to evaluate the variants of the $(1+1)$-EA and of the ($\mu+1$)-EA using these different operators for the dynamic \wdtsps{}. A key aspect when comparing the $(1+1)$-EA and ($\mu+1$)-EA is whether a larger population helps for dealing with the dynamic problem. Larger parent populations have the disadvantage of slowing down the optimization process for problems that can be solved quite easily by hill-climbing techniques. As the offspring in the ($\mu+1$)-EA only competes against its parent there is a smaller selection pressure than in another variant of this algorithms where after mutation always a worst individual out of the $\mu+1$ individuals is removed. We also considered this variant in our initial experimental investigations, but it performs significantly worse than our variant of the ($\mu+1$)-EA given in Algorithm~\ref{alg:ea}.

\section{Dynamic Setup}
\label{sec:sec4}

\ignore{
\begin{algorithm}[t]
\begin{algorithmic}
\STATE Choose initial bitstring $x^0 \in \{0,1\}^m$\\
\STATE t:=t+1\\
\IF{$t \mod \tau ==0$ }\\
    \STATE $y=x^t$.
    \STATE \FOR{i=1 to m}
    \STATE \IF{|x^t|_1>L and y_i==1} 
            \STATE flip y_i with probability $p_1 = r/|x^t|_1$
            \ENDIF
            \ENDFOR
            \ENDIF
  \caption{Dynamic Changes}
  \label{alg:ea}
  \end{algorithmic}
\end{algorithm}
}
\ignore{
\jakob{@Frank: could you please address the following comment raised by one of the AAAI-reviewers? \enquote{The proposed dynamic framework is not so motivated.The proposed dynamic framework is one of the contributions in this paper, and the authors in Section Conclusion point the easy applicability to other packing problems and expect this setup to be useful for other combinatorial problems. However, the proposed dynamic framework is not well-motivated. Although W-TSP is a recently proposed problem, I guess the dynamic setup is essentially performed on the packing component, and I would suggest to mention the existing dynamic frameworks on the general packing problems, is the proposed dynamic framework a totally new setup considering the packing problems? If yes, why do we need the new dynamic setup, and if not, we and readers are clearer about the novelty.}}}

We now describe a dynamic setting for \wdtsps. Recall that we denote by $x_{ij} = 1$ an active item $e_{ij}$, and by $x_{ij} = 0$ an inactive item $e_{ij}$. 
In the dynamic setting the bitstring $x \in \{0,1\}^m$ for a given instance changes with a given frequency denoted by $\tau$.

The number of items should be roughly within the interval $[L, U]$ where $L$ is the lower bound and $U$ the upper bound determined by the user. We denote $L$ and $U$ in terms of percentages of the number of $1$-bits that a bitstring $x$ contains. For example, $[70, 90]$ denotes the interval that contains all bitstrings having between $70\%$ and $90\%$ of $1$-bits. Note that the number of different bitstrings can be quite different even if the value of $(U-L)$ is the same. Consider for example the intervals $[50, 70]$ and $[70, 90]$. Clearly the first interval allows for significantly more different bitstrings than the second interval which is due to the binomial coefficients corresponding to the values in these intervals.

An important property of our approach is that it is focused on the number of bits set to $1$ and performs a random walk on the interval with respect to that number. To achieve this, we adapt the approach of flipping bits with asymmetric mutation probabilities. This approach has already been investigated in the context of runtime analysis~\cite{DBLP:journals/tcs/NeumannW07,DBLP:journals/ec/JansenS10} and shown to be beneficial for the creation of evolutionary image transition processes~\cite{DBLP:journals/ec/NeumannAN20}. It also resembles recent dynamic setups for problems with dynamic constraints where the constraint bound increase or decreases with equal probability~\cite{DBLP:conf/ppsn/Roostapour0N18,DBLP:journals/corr/abs-1811-07806}.

Let $x^t$ be the current bitstring determining the current set of available items. In order to obtain the set of available items after a dynamic change, we flip each $1$-bit and $0$-bit with probability 
\begin{align*}
    p^t_1 = \frac{r}{|x^t|_1}
\quad\text{and}\quad
    p^t_0 = \frac{r}{|x^t|_0}
\end{align*}
respectively to obtain the new bitstring $y$ where $|x^t|_k, k \in \{0,1\}$ is the number of $k$-bits in $x^t$. In order to make the dynamic setup dependent on the size of the problem (in terms of the total number of items $m$), we denote the magnitude of change as percentage $c$ of the overall number of items, i.e. we have 
$r = c \cdot m/100,$
where $c$ is the magnitude of change each of the categories of $1$-bits and $0$-bits. 

In our experimental investigations, we will always report the magnitude of change in terms of the percentage $c$.
Note that the total percentage of items that is expected to change is $2c$ as both the $1$-bit flips and the $0$-bit flips contribute $c$ to the total change.  However, the expected change in the number of $1$-bits is $0$ as in expectation the same number of $1$-bits and $0$-bits are flipped.

In order to make sure that the number of $1$-bits of $y$ can only leave interval $[L,U]$ by a large amount and return back to the interval afterwards, we refrain from carrying out $0$-bit flips or $1$-bit flips if the number of ones of $x^t$ is at or out of the corresponding interval boundary. More precisely, we only flip a $1$-bit of $x^t$ to obtain the new bitstring $y$ iff 
$|x^t|_1 > L\cdot m/100$ and only flip a $0$-bit of $x^t$ iff 
$|x^t|_0 < U\cdot m/100.$

Changes are carried out every $\tau$ fitness evaluations, where $\tau$ is the frequency of a given dynamic setup. This implies that every evolutionary algorithm considered in our study has $\tau$ iterations to compute a reoptimized solution after a dynamic change has occurred. The overall type of changes that are applied can therefore be specified by the parameter combination of $c$ and $\tau$ and the used interval $[L,U]$. In line with the literature on dynamic optimization we call the phase between two dynamic changes an \emph{epoch}.

Exemplary dynamic benchmarks showing $30$ changes are displayed in Figure~\ref{fig:dynamic_setup}. 
First, the Hamming-distance $HD(x^0, x^t)$ from the current bitstring $x^t$ to the bitstring $x^0$ from an initial packing plan on instance eil101 with $500$ items is shown in the top row. We obtain the results within the intervals $[30, 70]$ and $[70, 90]$. Subsequently, the magnitude of change $c \in \{2, 5, 10\}$ is visualized. 

The next row displays the Hamming-distance $HD(x^{i}, x^{i+1})$ between sequential bitstrings for $30$ dynamic changes. We can see that the number of bit positions in which the two sequential bitstrings are different increases as the magnitude of change increases (note the different scaling of the $y$-axis). This indicates that, as expected, the trajectory shows a stronger fluctuation while the Hamming-distance increases as the magnitude of change increases.

In the last row the number of active items $x_{ij} = 1$ 
for each dynamic change are displayed. We observe that the number of active items rise and fall both irregularly and differently for each presented setting. In conclusion,
we are able to obtain a trajectory of higher values for the number of active items within the intervals $[70, 90]$ in our dynamic setting.

\section{On the Impact of Dynamic Changes}
\label{sec:sec5}

\begin{figure*}[tb]
    \includegraphics[width=\textwidth, trim = 13pt 6pt 4pt 10pt, clip]{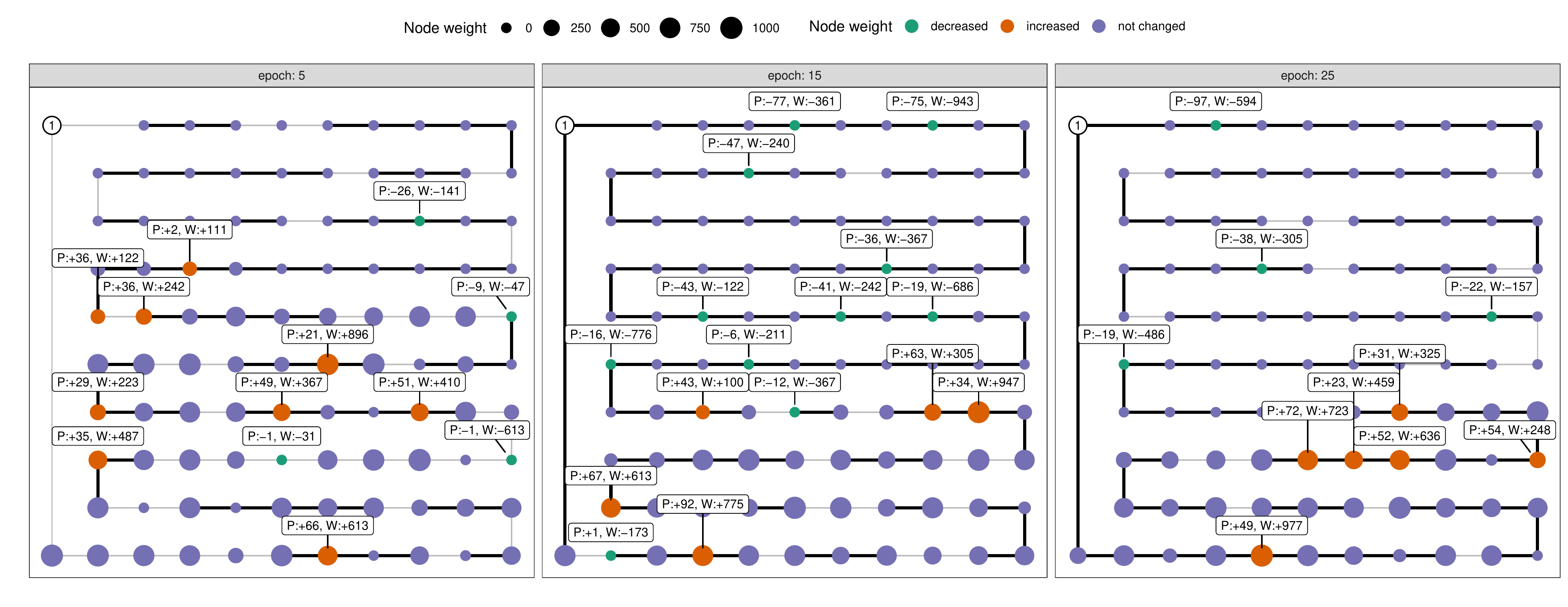}
    \includegraphics[width=\textwidth, trim=13pt 10pt 4pt 45pt, clip]{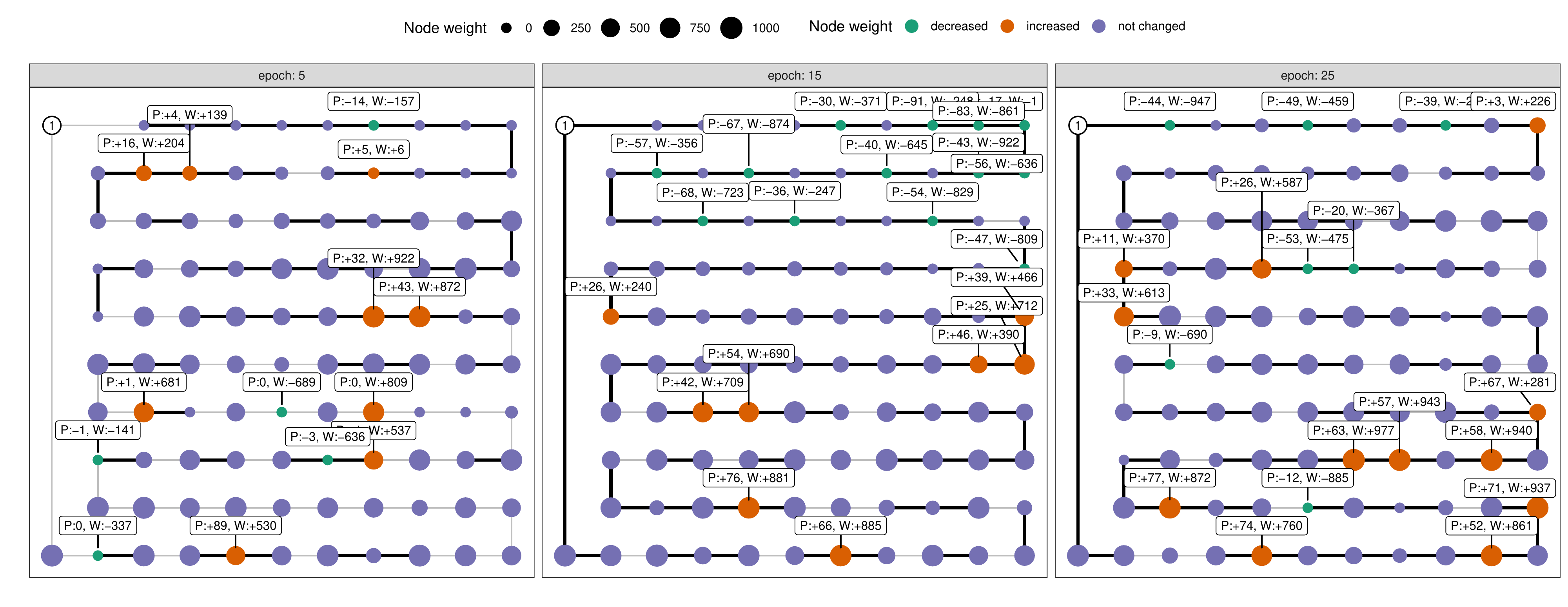}
    \caption{Visualization of best tour $\pi$ calculated by $(20+1)$-EA with inversion mutation in epochs 5, 15 and 25 (columns) with $\tau = 500\,000$ on instance eil101-uncorr with $100$ items. The dynamic changes are determined by $L=[30, 70], c=5$ (top row) and $L=[70,90], c=2$ (bottom row). Black edges are not part of the best tour in the last epoch. The size and color of nodes indicates if their weight was affected by the latest dynamic change; if so, labels indicate their replacement with respect to the initial tour (P) and the change in weight (W). For visual clarity nodes are positioned according to their position in the tour $\pi$. I.e., node 1 is positioned in the top left corner and the following nodes are arranged clockwise following their ordering in $\pi$.}
    \label{fig:dynamic_change_tour_affection}
\end{figure*}
We now carry out investigations to see how the changing weight on the nodes determined by the vector $x^t$ in the dynamic setting impacts changes to the recomputed tours. To do this, we use the $(20+1)$-EA where the offspring is created by inversion and set $\tau$ very large, $\tau=500,000$, in order to obtain highly optimised tours after the change.
Fig.~\ref{fig:dynamic_change_tour_affection} visualizes representative tours in epochs $5$, $15$ and $25$ respectively for instance eil101-uncorr with $m=100$ items. Since optimal tours for the \wdtsps{} in general are not necessarily free of crossings, for visual clarity and in order to be able to recognize patterns we arrange nodes with respect to their order in the permutation $\pi$.\footnote{In fact, embedding the eil101 instance with their Euclidean coordinates results in chaotic visualization of \wdtsps{} tours with many crossing edges. Here it is hard to spot changes and patterns.}
The plots allow for some interesting observations. There is a clear indication that heavy nodes are favorably visited later on while light nodes are visited earlier (see node size encoding in Fig.~\ref{fig:dynamic_change_tour_affection}). This is in line with intuition since placing heavy nodes in the beginning leads to an enormous accumulation of weight which is clearly disadvantageous in the \wdtsps{} setting (recall the \wdtsps{} fitness function in Eq.~(\ref{eq:wtsp_fitness_function})). Linked to this observation we see that nodes that were affected by the dynamic change (indicated by red and green points) tend to be rearranged to earlier positions in $\pi$ if their weight decreased and shifted to later positions if their weight increased. In the plots this is visualized by a negative $P$ (for position) value in the respective labels. For example, $P:-25$ means that the node was moved by $25$ positions to the front of the permutation in comparison to its position in the final solution of the previous epoch. There are few exceptions to this rule (see, e.g. the green node in the last row of the right plot). A reasonable justification is either that the algorithm run into a local optimum or, just as likely, there is a complex interplay between the node weights, the distances and the node placement in the Euclidean plane which affects the overall tour length.
Last but not least the number of new edges, i.e. edges that were not part of the initial solution to the respective epoch, is very high even in the considered setting with a low $c$-value (low percentage of changes). In consequence, reoptimization requires a lot of operations.
Our observations are consistent across a wide range of independent runs for different values of our dynamic setup.

\section{Experimental Comparison of Baseline Evolutionary Algorithms}
\label{sec:sec6}
%
We investigate two baseline evolutionary algorithms: the $(1+1)$-EA and the $(\mu+1)$-EA with $\mu=20$. We consider each algorithm in its three variants using inversion, exchange and jump as mutation operators. Our goal is to see which operators are best suited for dealing with our benchmark. Furthermore, we are interested in dynamic benchmark settings where the population of the $(\mu+1)$-EA gives a benefit over the $(1+1)$-EA. Note that the frequency $\tau$ gives the number of fitness evaluations that an algorithm can do between the dynamic changes. In the case that the problem is easy to reoptimize, a population might not be that helpful as it can slow down the reoptimization process.

Our benchmark set consists of benchmark instances eil101 and a280 from the TTP-benchmark set.\footnote{\url{https://cs.adelaide.edu.au/~optlog/TTP2017Comp/}} Furthermore, we consider each three categories of these instances, namely bounded strongly correlated (bcc), uncorrelated (uncorr) and uncorrelated with similar weights (usw).
Each problem instance has $m = 5\cdot (n-1)$ items, i.e. five items per node except for the first node. All combinations of $[L, U] \in \{[30, 70], [70,90]\}$, $c \in \{2, 5, 10\}$ and $\tau \in \{100,\linebreak[1] 1\,000,\linebreak[1] 10\,000,\linebreak[1] 50\,000,\linebreak[1] 75\,000,\linebreak[1] 100\,000,\linebreak[1] 250\,000,\linebreak[1] 500\,000,\linebreak[1] 750\,000\}$ are tested and for each $([L, U], c, \tau)$-combination in total 30 different packing plan sequences are generated. The number of epochs is $31$ (including a 0-epoch before the packing plan is altered for the first time). In the 0-epoch, we run the algorithms for $50\,000$ function evaluations independently of $\tau$ to obtain a good starting point.

\begin{table*}[htbp]
\caption{\label{tab:all_results}Mean (\textbf{mean}), standard deviation (\textbf{std}) and results of Wilcoxon-Mann-Whitney tests at a significance level of $\alpha = 0.05$ (\textbf{stat}) in terms of relative performance for all considered instances. Best, i.e. lowest, mean values are highlighted in \colorbox{gray!20}{\textbf{bold face}}. The entries in the stat-column are read as follows: a value of $S \in \{1, \ldots, 6\}$ means that algorithm $S$ performs significantly worse. Interval notation $S-T$ means that all algorithms in the interval perform worse.}
\renewcommand{\arraystretch}{0.7}
\renewcommand{\tabcolsep}{4pt}
\centering
\begin{tiny}
\begin{tabular}[t]{crrrrrrrrrrrrrrrrrrrrr}
\toprule
\multicolumn{1}{c}{\textbf{ }} & \multicolumn{1}{c}{\textbf{ }} & \multicolumn{1}{c}{\textbf{ }} & \multicolumn{1}{c}{\textbf{ }} & \multicolumn{9}{c}{\textbf{(1+1) EA}} & \multicolumn{9}{c}{\textbf{(20+1)-EA}} \\
\cmidrule(l{3pt}r{3pt}){5-13} \cmidrule(l{3pt}r{3pt}){14-22}
\multicolumn{1}{c}{\textbf{ }} & \multicolumn{1}{c}{\textbf{ }} & \multicolumn{1}{c}{\textbf{ }} & \multicolumn{1}{c}{\textbf{ }} & \multicolumn{3}{c}{\textbf{Inversion (1)}} & \multicolumn{3}{c}{\textbf{Swap (2)}} & \multicolumn{3}{c}{\textbf{Jump (3)}} & \multicolumn{3}{c}{\textbf{Inversion (4)}} & \multicolumn{3}{c}{\textbf{Swap (5)}} & \multicolumn{3}{c}{\textbf{Jump (6)}} \\
\cmidrule(l{3pt}r{3pt}){5-7} \cmidrule(l{3pt}r{3pt}){8-10} \cmidrule(l{3pt}r{3pt}){11-13} \cmidrule(l{3pt}r{3pt}){14-16} \cmidrule(l{3pt}r{3pt}){17-19} \cmidrule(l{3pt}r{3pt}){20-22}
& $\tau$ & $[L, U]$ & $c$ & \textbf{mean} & \textbf{std} & \textbf{stat} & \textbf{mean} & \textbf{std} & \textbf{stat} & \textbf{mean} & \textbf{std} & \textbf{stat} & \textbf{mean} & \textbf{std} & \textbf{stat} & \textbf{mean} & \textbf{std} & \textbf{stat} & \textbf{mean} & \textbf{std} & \textbf{stat}\\
\midrule
 &  &  & 2 & 22.8 & 9.5 & 3-6 & \cellcolor{gray!20}{\textbf{16.0}} & 7.3 & 1,3-6 & 38.0 & 11.2 &  & 39.7 & 30.0 &  & 40.0 & 19.2 &  & 35.7 & 17.7 & 4-5\\

 &  &  & 5 & 18.8 & 9.0 & 3-6 & \cellcolor{gray!20}{\textbf{15.6}} & 7.6 & 3-6 & 43.8 & 14.3 &  & 41.6 & 29.6 & 5 & 44.9 & 18.7 &  & 38.6 & 18.3 & 3,5\\

 &  & \multirow{-3}{*}{\raggedleft\arraybackslash $[30, 70]$} & 10 & 16.7 & 10.0 & 3-6 & \cellcolor{gray!20}{\textbf{15.3}} & 8.1 & 3-6 & 49.2 & 16.7 &  & 45.8 & 30.5 & 3,5 & 50.2 & 19.2 &  & 42.6 & 19.0 & 3-5\\

\cmidrule{3-22}
 &  &  & 2 & \cellcolor{gray!20}{\textbf{19.1}} & 6.7 & 3-6 & 21.8 & 6.1 & 3-6 & 35.3 & 7.8 &  & 34.3 & 30.4 & 5 & 40.2 & 20.0 &  & 30.0 & 19.4 & 3-5\\

 &  &  & 5 & \cellcolor{gray!20}{\textbf{13.3}} & 5.6 & 2-6 & 19.2 & 6.6 & 3-6 & 31.5 & 8.4 & 5 & 34.1 & 29.7 & 5 & 41.1 & 20.4 &  & 29.1 & 19.4 & 3-5\\

 & \multirow{-6}{*}{\raggedleft\arraybackslash \rotatebox{90}{$10\,000$}} & \multirow{-3}{*}{\raggedleft\arraybackslash $[70, 90]$} & 10 & \cellcolor{gray!20}{\textbf{11.5}} & 6.1 & 2-6 & 19.2 & 7.3 & 3-6 & 29.5 & 10.1 & 5 & 33.7 & 30.1 & 5 & 42.0 & 21.1 &  & 32.7 & 19.5 & 5\\

\cmidrule{2-22}
 &  &  & 2 & 20.8 & 8.1 & 3 & 15.6 & 7.1 & 3 & 38.2 & 11.4 &  & 3.0 & 10.8 & 1-3,5 & 11.0 & 9.3 & 2-3 & \cellcolor{gray!20}{\textbf{-0.3}} & 12.4 & 1-5\\

 &  &  & 5 & 15.9 & 8.7 & 3 & 14.0 & 7.5 & 3 & 38.8 & 13.1 &  & 3.6 & 11.1 & 1-3,5 & 12.3 & 9.6 & 3 & \cellcolor{gray!20}{\textbf{1.5}} & 10.7 & 1-5\\

 &  & \multirow{-3}{*}{\raggedleft\arraybackslash $[30, 70]$} & 10 & 15.4 & 9.9 & 3 & 9.3 & 7.3 & 3,5 & 42.7 & 16.4 &  & \cellcolor{gray!20}{\textbf{4.6}} & 11.3 & 1-3,5 & 12.7 & 9.9 & 3 & 5.5 & 10.2 & 1-3,5\\

\cmidrule{3-22}
 &  &  & 2 & 14.2 & 5.8 & 2-3 & 21.1 & 5.9 & 3 & 33.4 & 7.3 &  & 1.7 & 10.6 & 1-3,5 & 13.9 & 9.4 & 2-3 & \cellcolor{gray!20}{\textbf{-0.9}} & 11.9 & 1-5\\

 &  &  & 5 & 13.8 & 6.3 & 3 & 14.4 & 6.4 & 3 & 29.3 & 7.7 &  & 0.6 & 10.4 & 1-3,5 & 12.9 & 9.4 & 3 & \cellcolor{gray!20}{\textbf{-0.6}} & 11.4 & 1-5\\

\multirow{-14}{*}{\centering\arraybackslash \rotatebox{90}{\textbf{eil101-bsc}}} & \multirow{-6}{*}{\raggedleft\arraybackslash \rotatebox{90}{$75\,000$}} & \multirow{-3}{*}{\raggedleft\arraybackslash $[70, 90]$} & 10 & 13.2 & 6.0 & 3 & 16.2 & 6.2 & 3 & 28.4 & 8.5 &  & 0.8 & 10.9 & 1-3,5 & 12.4 & 9.9 & 2-3 & \cellcolor{gray!20}{\textbf{0.2}} & 11.6 & 1-3,5\\
\cmidrule{1-22}
 &  &  & 2 & \cellcolor{gray!20}{\textbf{15.4}} & 7.0 & 2-6 & 21.5 & 7.0 & 3-6 & 41.2 & 10.7 &  & 38.1 & 29.3 & 5 & 41.7 & 19.0 &  & 32.5 & 18.4 & 3-5\\

 &  &  & 5 & \cellcolor{gray!20}{\textbf{16.8}} & 8.7 & 3-6 & 17.9 & 8.1 & 3-6 & 42.8 & 12.1 &  & 39.6 & 29.4 & 3,5 & 46.3 & 19.4 &  & 36.7 & 18.2 & 3-5\\

 &  & \multirow{-3}{*}{\raggedleft\arraybackslash $[30, 70]$} & 10 & \cellcolor{gray!20}{\textbf{15.1}} & 7.3 & 3-6 & 16.0 & 7.7 & 3-6 & 45.2 & 15.4 &  & 43.8 & 28.7 & 5 & 50.9 & 18.5 &  & 40.4 & 18.3 & 4-5\\

\cmidrule{3-22}
 &  &  & 2 & \cellcolor{gray!20}{\textbf{14.8}} & 4.9 & 3-6 & 19.6 & 5.6 & 3-6 & 32.6 & 6.7 & 5 & 32.3 & 29.6 & 5 & 42.6 & 20.9 &  & 29.9 & 18.9 & 4-5\\

 &  &  & 5 & \cellcolor{gray!20}{\textbf{11.7}} & 5.7 & 2-6 & 21.9 & 6.5 & 3-6 & 27.3 & 7.9 & 4-6 & 32.3 & 29.0 & 5 & 44.3 & 21.6 &  & 30.3 & 19.1 & 4-5\\

 & \multirow{-6}{*}{\raggedleft\arraybackslash \rotatebox{90}{$10\,000$}} & \multirow{-3}{*}{\raggedleft\arraybackslash $[70, 90]$} & 10 & \cellcolor{gray!20}{\textbf{8.4}} & 6.8 & 2-6 & 18.7 & 7.3 & 3-6 & 29.7 & 8.4 & 5 & 33.4 & 29.9 & 5 & 43.9 & 21.4 &  & 31.4 & 18.7 & 4-5\\

\cmidrule{2-22}
 &  &  & 2 & 17.8 & 7.5 & 3 & 15.8 & 6.9 & 3 & 35.7 & 10.2 &  & 2.8 & 10.3 & 1-3,5 & 12.5 & 9.5 & 3 & \cellcolor{gray!20}{\textbf{0.6}} & 12.0 & 1-5\\

 &  &  & 5 & 10.2 & 7.8 & 2-3,5 & 13.1 & 7.1 & 3 & 38.0 & 12.9 &  & 3.4 & 10.9 & 2-3,5 & 13.8 & 10.1 & 3 & \cellcolor{gray!20}{\textbf{2.6}} & 11.2 & 2-5\\

 &  & \multirow{-3}{*}{\raggedleft\arraybackslash $[30, 70]$} & 10 & 14.3 & 8.1 & 3 & 10.9 & 6.8 & 3,5 & 39.3 & 15.3 &  & \cellcolor{gray!20}{\textbf{4.4}} & 10.6 & 1-3,5-6 & 13.4 & 9.4 & 3 & 5.5 & 10.0 & 1-3,5\\

\cmidrule{3-22}
 &  &  & 2 & 12.7 & 5.2 & 2-3 & 22.4 & 5.3 & 3 & 31.0 & 6.2 &  & 1.5 & 10.6 & 1-3,5 & 15.1 & 9.5 & 2-3 & \cellcolor{gray!20}{\textbf{0.1}} & 11.2 & 1-5\\

 &  &  & 5 & 10.7 & 5.5 & 2-3,5 & 17.2 & 5.8 & 3 & 27.6 & 7.3 &  & 1.0 & 10.6 & 1-3,5 & 14.9 & 10.1 & 3 & \cellcolor{gray!20}{\textbf{0.8}} & 10.8 & 1-3,5\\

\multirow{-14}{*}{\centering\arraybackslash \rotatebox{90}{\textbf{eil101-uncorr}}} & \multirow{-6}{*}{\raggedleft\arraybackslash \rotatebox{90}{$75\,000$}} & \multirow{-3}{*}{\raggedleft\arraybackslash $[70, 90]$} & 10 & 8.8 & 5.2 & 2-3,5 & 17.1 & 6.4 & 3 & 26.4 & 7.6 &  & 0.9 & 10.8 & 1-3,5 & 13.6 & 10.1 & 2-3 & 0.9 & 11.0 & 1-3,5\\
\cmidrule{1-22}
 &  &  & 2 & \cellcolor{gray!20}{\textbf{17.2}} & 7.1 & 2-6 & 22.4 & 6.8 & 3-6 & 35.4 & 8.7 & 5 & 36.1 & 29.0 & 5 & 43.6 & 18.6 &  & 31.8 & 18.6 & 3-5\\

 &  &  & 5 & \cellcolor{gray!20}{\textbf{15.6}} & 6.9 & 2-6 & 20.4 & 7.2 & 3-6 & 37.1 & 11.3 & 5 & 38.3 & 30.1 & 5 & 46.8 & 19.4 &  & 34.8 & 19.1 & 4-5\\

 &  & \multirow{-3}{*}{\raggedleft\arraybackslash $[30, 70]$} & 10 & \cellcolor{gray!20}{\textbf{16.1}} & 7.8 & 3-6 & 17.9 & 7.8 & 3-6 & 43.2 & 12.9 &  & 39.8 & 30.6 & 3,5 & 49.8 & 19.9 &  & 36.5 & 19.3 & 3-5\\

\cmidrule{3-22}
 &  &  & 2 & \cellcolor{gray!20}{\textbf{12.2}} & 5.2 & 2-6 & 22.5 & 6.1 & 3-6 & 26.7 & 6.7 & 5 & 32.2 & 29.5 & 5 & 43.4 & 21.0 &  & 29.6 & 19.2 & 4-5\\

 &  &  & 5 & \cellcolor{gray!20}{\textbf{12.4}} & 5.5 & 2-6 & 21.9 & 6.1 & 3-6 & 26.6 & 7.3 & 4-5 & 33.2 & 29.4 & 5 & 44.4 & 21.3 &  & 31.8 & 20.0 & 5\\

 & \multirow{-6}{*}{\raggedleft\arraybackslash \rotatebox{90}{$10\,000$}} & \multirow{-3}{*}{\raggedleft\arraybackslash $[70, 90]$} & 10 & \cellcolor{gray!20}{\textbf{8.1}} & 5.0 & 2-6 & 20.0 & 6.1 & 3-6 & 27.4 & 7.2 & 4-5 & 33.2 & 29.1 & 5 & 43.5 & 21.2 &  & 29.5 & 19.3 & 4-5\\

\cmidrule{2-22}
 &  &  & 2 & 12.8 & 6.9 & 2-3 & 19.7 & 6.4 & 3 & 34.5 & 7.6 &  & 2.3 & 10.4 & 1-3,5 & 14.0 & 9.1 & 2-3 & \cellcolor{gray!20}{\textbf{0.4}} & 11.1 & 1-5\\

 &  &  & 5 & 13.7 & 7.2 & 2-3 & 17.8 & 7.3 & 3 & 31.4 & 10.2 &  & 2.5 & 10.6 & 1-3,5 & 14.8 & 9.7 & 2-3 & \cellcolor{gray!20}{\textbf{2.1}} & 10.7 & 1-3,5\\

 &  & \multirow{-3}{*}{\raggedleft\arraybackslash $[30, 70]$} & 10 & 14.1 & 7.6 & 3 & 12.9 & 7.1 & 3,5 & 35.8 & 13.2 &  & \cellcolor{gray!20}{\textbf{3.1}} & 11.4 & 1-3,5 & 14.2 & 10.2 & 3 & 3.9 & 10.9 & 1-3,5\\

\cmidrule{3-22}
 &  &  & 2 & 13.3 & 5.4 & 2-3,5 & 22.9 & 5.4 &  & 26.3 & 6.8 &  & 1.2 & 10.4 & 1-3,5 & 16.4 & 9.1 & 2-3 & \cellcolor{gray!20}{\textbf{1.0}} & 10.5 & 1-3,5\\

 &  &  & 5 & 13.5 & 4.9 & 2-3 & 19.6 & 5.5 & 3 & 26.8 & 7.1 &  & 1.4 & 10.5 & 1-3,5 & 15.6 & 9.8 & 2-3 & \cellcolor{gray!20}{\textbf{0.8}} & 11.1 & 1-3,5\\

\multirow{-14}{*}{\centering\arraybackslash \rotatebox{90}{\textbf{eil101-usw}}} & \multirow{-6}{*}{\raggedleft\arraybackslash \rotatebox{90}{$75\,000$}} & \multirow{-3}{*}{\raggedleft\arraybackslash $[70, 90]$} & 10 & 11.6 & 5.1 & 2-3 & 18.6 & 5.7 & 3 & 23.8 & 7.4 &  & \cellcolor{gray!20}{\textbf{0.8}} & 10.1 & 1-3,5 & 15.0 & 9.7 & 2-3 & 1.3 & 9.8 & 1-3,5\\
\midrule
 &  &  & 2 & \cellcolor{gray!20}{\textbf{18.3}} & 9.4 & 2-6 & 51.7 & 12.3 & 3,5 & 86.8 & 15.9 &  & 52.9 & 54.9 & 3,5 & 84.0 & 31.0 & 3 & 49.8 & 37.8 & 3,5\\

 &  &  & 5 & \cellcolor{gray!20}{\textbf{18.7}} & 9.8 & 2-6 & 44.8 & 13.1 & 3-6 & 95.2 & 20.5 &  & 55.1 & 53.7 & 3,5 & 87.4 & 30.5 & 3 & 52.4 & 38.0 & 3,5\\

 &  & \multirow{-3}{*}{\raggedleft\arraybackslash $[30, 70]$} & 10 & \cellcolor{gray!20}{\textbf{17.9}} & 10.1 & 2-6 & 41.3 & 14.3 & 3-6 & 96.1 & 26.6 &  & 56.1 & 55.2 & 3,5 & 91.1 & 30.7 & 3 & 52.0 & 37.5 & 3-5\\

\cmidrule{3-22}
 &  &  & 2 & \cellcolor{gray!20}{\textbf{17.1}} & 8.5 & 2-6 & 68.2 & 12.3 & 3,5 & 78.6 & 14.6 & 5 & 49.1 & 55.8 & 2-3,5 & 95.5 & 35.1 &  & 49.0 & 38.1 & 2-3,5\\

 &  &  & 5 & \cellcolor{gray!20}{\textbf{14.7}} & 6.9 & 2-6 & 61.4 & 13.2 & 3,5 & 65.9 & 15.2 & 5 & 48.9 & 56.3 & 2-3,5 & 97.4 & 35.1 &  & 52.1 & 38.4 & 2-3,5\\

 & \multirow{-6}{*}{\raggedleft\arraybackslash \rotatebox{90}{100\,000}} & \multirow{-3}{*}{\raggedleft\arraybackslash $[70, 90]$} & 10 & \cellcolor{gray!20}{\textbf{16.4}} & 8.1 & 2-6 & 55.5 & 14.2 & 3,5 & 69.1 & 18.8 & 5 & 49.5 & 55.5 & 2-3,5 & 95.6 & 34.4 &  & 44.6 & 36.0 & 2-5\\

\cmidrule{2-22}
 &  &  & 2 & 19.7 & 9.7 & 2-3,5 & 46.6 & 10.5 & 3 & 83.1 & 18.3 &  & 2.4 & 13.7 & 1-3,5 & 41.2 & 14.4 & 2-3 & \cellcolor{gray!20}{\textbf{-3.5}} & 17.1 & 1-5\\

 &  &  & 5 & 9.1 & 8.1 & 2-3,5 & 38.5 & 11.9 & 3 & 85.5 & 23.7 &  & 2.4 & 13.9 & 2-3,5 & 38.6 & 15.6 & 3 & \cellcolor{gray!20}{\textbf{0.7}} & 18.8 & 1-5\\

 &  & \multirow{-3}{*}{\raggedleft\arraybackslash $[30, 70]$} & 10 & 11.2 & 8.1 & 2-3,5 & 34.7 & 12.8 & 3,5 & 73.0 & 29.2 &  & \cellcolor{gray!20}{\textbf{1.5}} & 14.6 & 2-3,5 & 38.3 & 16.4 & 3 & 2.6 & 17.0 & 2-3,5\\

\cmidrule{3-22}
 &  &  & 2 & 16.5 & 6.8 & 2-3,5 & 61.5 & 9.5 & 3 & 65.7 & 12.7 &  & 2.2 & 13.9 & 1-3,5 & 54.5 & 14.9 & 2-3 & \cellcolor{gray!20}{\textbf{-3.6}} & 15.9 & 1-5\\

 &  &  & 5 & 11.7 & 6.3 & 2-3,5 & 59.2 & 10.3 &  & 58.7 & 13.5 &  & 0.9 & 14.0 & 1-3,5 & 53.0 & 15.7 & 2 & \cellcolor{gray!20}{\textbf{-2.4}} & 17.6 & 1-5\\

\multirow{-14}{*}{\centering\arraybackslash \rotatebox{90}{\textbf{a280-bsc}}} & \multirow{-6}{*}{\raggedleft\arraybackslash \rotatebox{90}{750\,000}} & \multirow{-3}{*}{\raggedleft\arraybackslash $[70, 90]$} & 10 & 17.4 & 6.3 & 2-3,5 & 49.8 & 12.1 & 3 & 64.4 & 16.1 &  & 0.0 & 14.3 & 1-3,5 & 48.5 & 16.6 & 3 & \cellcolor{gray!20}{\textbf{-1.9}} & 17.4 & 1-5\\
\cmidrule{1-22}
 &  &  & 2 & \cellcolor{gray!20}{\textbf{20.0}} & 10.1 & 2-6 & 57.8 & 13.2 & 3,5 & 103.2 & 15.8 &  & 52.5 & 54.2 & 2-3,5 & 92.0 & 32.3 & 3 & 50.8 & 38.9 & 2-3,5\\

 &  &  & 5 & \cellcolor{gray!20}{\textbf{14.2}} & 10.1 & 2-6 & 49.3 & 14.7 & 3-5 & 96.4 & 20.9 &  & 53.6 & 53.5 & 3,5 & 95.3 & 31.2 &  & 51.6 & 38.4 & 3,5\\

 &  & \multirow{-3}{*}{\raggedleft\arraybackslash $[30, 70]$} & 10 & \cellcolor{gray!20}{\textbf{11.5}} & 9.6 & 2-6 & 48.4 & 15.7 & 3-5 & 94.4 & 28.0 &  & 54.3 & 56.3 & 3,5 & 99.1 & 33.9 &  & 51.0 & 37.1 & 3,5\\

\cmidrule{3-22}
 &  &  & 2 & \cellcolor{gray!20}{\textbf{14.4}} & 7.3 & 2-6 & 73.5 & 11.9 & 5 & 69.9 & 14.7 & 5 & 49.1 & 55.6 & 2-3,5 & 105.1 & 35.8 &  & 47.8 & 37.2 & 2-5\\

 &  &  & 5 & \cellcolor{gray!20}{\textbf{11.2}} & 7.3 & 2-6 & 64.2 & 14.2 & 5 & 66.7 & 13.3 & 5 & 48.9 & 55.9 & 2-3,5 & 104.2 & 36.3 &  & 48.3 & 38.3 & 2-3,5\\

 & \multirow{-6}{*}{\raggedleft\arraybackslash \rotatebox{90}{100\,000}} & \multirow{-3}{*}{\raggedleft\arraybackslash $[70, 90]$} & 10 & \cellcolor{gray!20}{\textbf{10.4}} & 8.2 & 2-6 & 63.1 & 15.3 & 5 & 54.0 & 18.6 & 5 & 49.2 & 56.5 & 2,5 & 104.7 & 36.8 &  & 48.2 & 35.6 & 2,5\\

\cmidrule{2-22}
 &  &  & 2 & 12.7 & 7.6 & 2-3,5 & 53.0 & 11.0 & 3 & 85.5 & 17.0 &  & 2.4 & 13.6 & 1-3,5 & 47.5 & 14.6 & 2-3 & \cellcolor{gray!20}{\textbf{-0.9}} & 17.8 & 1-5\\

 &  &  & 5 & 11.6 & 7.9 & 2-3,5 & 43.1 & 12.7 & 3 & 88.1 & 19.9 &  & 1.5 & 13.8 & 1-3,5 & 44.9 & 16.0 & 3 & \cellcolor{gray!20}{\textbf{0.4}} & 15.9 & 1-5\\

 &  & \multirow{-3}{*}{\raggedleft\arraybackslash $[30, 70]$} & 10 & 7.4 & 9.9 & 2-3,5 & 38.1 & 14.3 & 3,5 & 86.9 & 24.4 &  & \cellcolor{gray!20}{\textbf{0.8}} & 14.8 & 1-3,5-6 & 43.7 & 18.1 & 3 & 3.4 & 17.3 & 2-3,5\\

\cmidrule{3-22}
 &  &  & 2 & 13.9 & 6.2 & 2-3,5 & 68.9 & 10.4 &  & 57.2 & 12.2 &  & 1.8 & 13.2 & 1-3,5 & 60.4 & 15.6 & 2 & \cellcolor{gray!20}{\textbf{-0.2}} & 18.2 & 1-5\\

 &  &  & 5 & 13.0 & 6.1 & 2-3,5 & 64.1 & 11.9 &  & 52.2 & 14.0 & 2 & -0.1 & 14.5 & 1-3,5 & 57.9 & 17.2 & 2 & \cellcolor{gray!20}{\textbf{-0.9}} & 17.4 & 1-3,5\\

\multirow{-14}{*}{\centering\arraybackslash \rotatebox{90}{\textbf{a280-uncorr}}} & \multirow{-6}{*}{\raggedleft\arraybackslash \rotatebox{90}{750\,000}} & \multirow{-3}{*}{\raggedleft\arraybackslash $[70, 90]$} & 10 & 15.6 & 6.4 & 2-3,5 & 54.0 & 12.8 &  & 53.4 & 14.1 &  & \cellcolor{gray!20}{\textbf{0.2}} & 14.4 & 1-3,5 & 55.5 & 17.6 &  & 0.6 & 18.2 & 1-3,5\\
\cmidrule{1-22}
 &  &  & 2 & \cellcolor{gray!20}{\textbf{12.1}} & 8.2 & 2-6 & 66.4 & 13.4 & 3,5 & 91.6 & 14.1 &  & 51.7 & 55.5 & 2-3,5 & 99.1 & 34.2 &  & 49.4 & 36.7 & 2-5\\

 &  &  & 5 & \cellcolor{gray!20}{\textbf{19.0}} & 9.2 & 2-6 & 57.2 & 14.8 & 3,5 & 95.4 & 20.0 &  & 53.3 & 55.8 & 3,5 & 100.4 & 33.9 &  & 50.3 & 36.4 & 2-5\\

 &  & \multirow{-3}{*}{\raggedleft\arraybackslash $[30, 70]$} & 10 & \cellcolor{gray!20}{\textbf{7.9}} & 10.2 & 2-6 & 52.1 & 16.2 & 3,5 & 78.5 & 25.0 & 5 & 52.7 & 56.6 & 3,5 & 104.6 & 35.1 &  & 52.1 & 39.4 & 3,5\\

\cmidrule{3-22}
 &  &  & 2 & \cellcolor{gray!20}{\textbf{14.6}} & 7.9 & 2-6 & 71.8 & 12.2 & 5 & 63.0 & 12.0 & 2,5 & 50.9 & 56.3 & 2-3,5 & 110.4 & 37.5 &  & 49.6 & 36.2 & 2-3,5\\

 &  &  & 5 & \cellcolor{gray!20}{\textbf{15.2}} & 7.6 & 2-6 & 65.8 & 14.1 & 5 & 55.7 & 13.9 & 5 & 50.1 & 57.6 & 2,5 & 111.5 & 38.6 &  & 52.0 & 35.5 & 2,5\\

 & \multirow{-6}{*}{\raggedleft\arraybackslash \rotatebox{90}{100\,000}} & \multirow{-3}{*}{\raggedleft\arraybackslash $[70, 90]$} & 10 & \cellcolor{gray!20}{\textbf{14.1}} & 7.9 & 2-6 & 64.1 & 15.7 & 5 & 63.3 & 15.3 & 5 & 50.3 & 57.4 & 2-3,5 & 111.3 & 38.2 &  & 51.3 & 36.3 & 2-3,5\\

\cmidrule{2-22}
 &  &  & 2 & 16.3 & 6.2 & 2-3,5 & 56.0 & 11.5 & 3 & 86.8 & 14.0 &  & 2.2 & 14.0 & 1-3,5 & 53.3 & 15.7 & 3 & \cellcolor{gray!20}{\textbf{-0.2}} & 19.5 & 1-5\\

 &  &  & 5 & 9.1 & 7.2 & 2-3,5 & 45.5 & 12.4 & 3,5 & 79.7 & 23.1 &  & 1.4 & 15.0 & 1-3,5 & 49.0 & 17.3 & 3 & \cellcolor{gray!20}{\textbf{0.7}} & 17.2 & 1-3,5\\

 &  & \multirow{-3}{*}{\raggedleft\arraybackslash $[30, 70]$} & 10 & 7.8 & 9.0 & 2-3,5 & 44.4 & 13.3 & 3,5 & 76.7 & 23.2 &  & \cellcolor{gray!20}{\textbf{0.2}} & 15.3 & 2-3,5-6 & 48.9 & 18.4 & 3 & 3.2 & 17.9 & 2-3,5\\

\cmidrule{3-22}
 &  &  & 2 & 15.0 & 5.1 & 2-3,5 & 67.9 & 9.8 &  & 64.4 & 10.3 &  & 2.3 & 14.0 & 1-3,5 & 62.9 & 16.9 & 2 & \cellcolor{gray!20}{\textbf{1.2}} & 17.5 & 1-5\\

 &  &  & 5 & 16.9 & 5.4 & 2-3,5 & 63.7 & 11.2 &  & 49.0 & 11.3 & 2,5 & 1.8 & 14.6 & 1-3,5 & 61.5 & 18.2 &  & \cellcolor{gray!20}{\textbf{1.3}} & 17.6 & 1-3,5\\

\multirow{-14}{*}{\centering\arraybackslash \rotatebox{90}{\textbf{a280-usw}}} & \multirow{-6}{*}{\raggedleft\arraybackslash \rotatebox{90}{750\,000}} & \multirow{-3}{*}{\raggedleft\arraybackslash $[70, 90]$} & 10 & 12.2 & 5.4 & 2-3,5 & 59.1 & 12.0 &  & 43.3 & 10.1 & 2,5 & \cellcolor{gray!20}{\textbf{0.7}} & 15.5 & 1-3,5 & 58.0 & 19.1 &  & 1.1 & 17.5 & 1-3,5\\
\bottomrule
\end{tabular}
\end{tiny}
\end{table*}

\begin{figure*}[!tb]
    \centering
    \includegraphics[width=\textwidth, trim=5pt 7pt 5pt 10pt, clip]{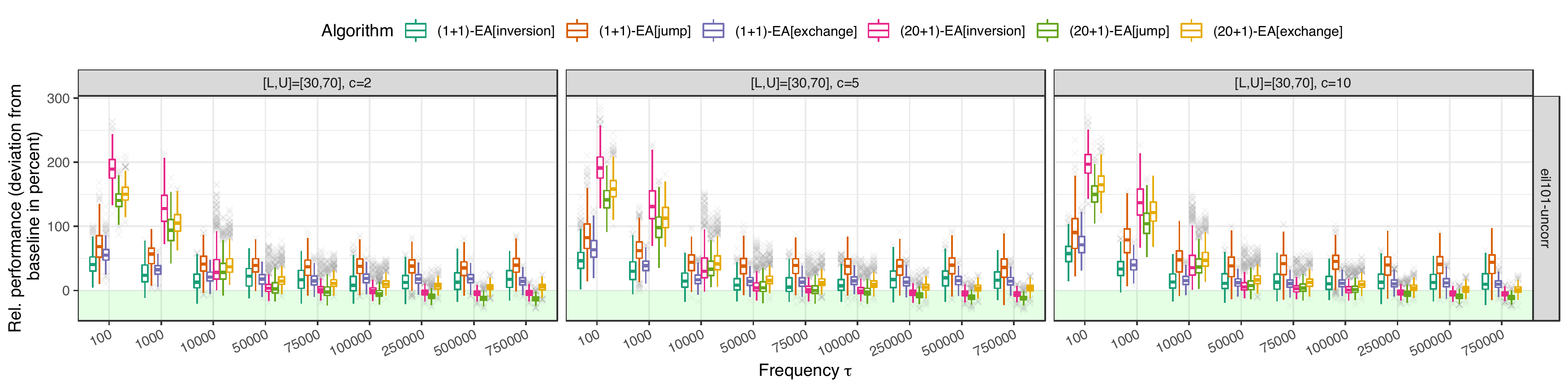}
    \caption{Distribution of performance by means of the deviation from the baseline in percent. Data is shown for problem instance eil101-uncorr and $[L=30, U=70]$, but the overall picture is the same across all instances and configurations. The green area below zero indicates the region where the solution quality obtained by the dynamic approach is better than the baseline obtained.}
    \label{fig:boxplots_deviation_distribution}
\end{figure*}
In addition, we treat each single packing plan as a static/offline problem and run $(20+1)$-EA with inversion mutation each 10 times for a total of $1\,000\,000$ function evaluations, i.e. longer than the maximum $\tau$ value in our setup. This is done to obtain a baseline approximate tour length for performance measurement. Assume the packing plan fixed (this includes parameters $L$, $U$ and $c$) and consider an epoch $d \in \{0, 30\}$. Let $\pi^{*}$ denote the shortest \wdtsps{} tour found by the offline approach and $\pi$ be the tour obtained by one of the six considered algorithms in our portfolio given $\tau$ fixed in epoch $d$.
We calculate the relative performance 
\begin{align*}
\text{perf}(\pi) = \left(\frac{\wtour(\pi)}{\wtour(\pi^{*})} - 1\right) \cdot 100.
\end{align*}
These values can be interpreted as the percentage of overshooting the baseline. Note though, that negative values are possible. This happens when the dynamic approach manages to find solutions that are superior to the baseline.
Note that ideally, we would like to compare to optimal solutions $\pi^{\text{OPT}}$. However, no exact algorithms are known so far that manage to find optimal \wdtsps{}-solutions for the instance sizes considered here in reasonable time (in contrast to, e.g., Concorde~\cite{applegate2007} for the classical TSP problem which manages to find optimal solutions for instances with hundreds of nodes within seconds~\cite{cook2012}). 

All experiments were conducted in the statistical programming language R~\cite{Rlang} with critical components written in C++. The package \texttt{batchtools}~\cite{Rbatchtools} was used to manage parallel jobs on a HPC cluster with Intel X86-64 Haswell CPUs.
For reproducibility we provide the implementations and experimental scripts\footnote{Data and code: \url{https://github.com/jakobbossek/GECCO2023-Dynamic-WTSP}} of our algorithms in the R package \texttt{TTP}\footnote{TTP:
\url{https://github.com/jakobbossek/TTP}}.
x

\subsection{Evaluation of Experimental Results}

In Table~\ref{tab:all_results}, we provide statistics on the relative performance of the eil101 and a280 instances with $5$ items per node for the final populations of $(1+1)$-EA and $(20+1)$-EA, with inversion, swap, and jump respectively acting as mutation operators.

For each algorithm and mutation operator the mean and standard deviation of the relative performance value of the final population of $30$ runs is shown. For the eil101 instances we report on different frequencies $10000, 75000$, the magnitude of changes $c = 2, 5, 10$ and the intervals $[30, 70]$ and $[70, 90]$. For the a280 instances we show $\tau \in \{100\,000,750\,000\}$. Note that due to space limitations we cannot report on all $\tau$-values in the result tables.
\begin{figure*}[htb]
    \centering
    \includegraphics[width=\textwidth, trim=0 16pt 0 0pt,clip]{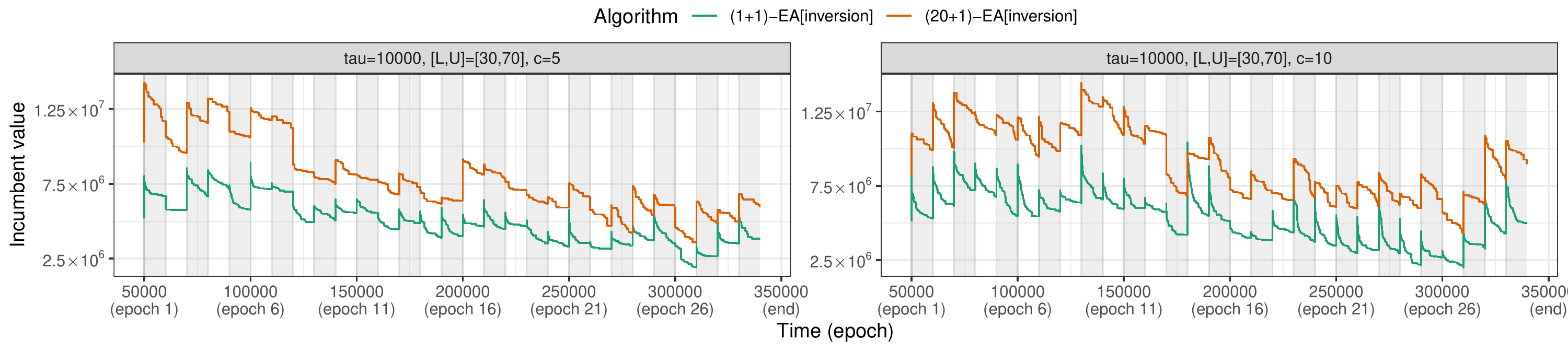}
    \includegraphics[width=\textwidth, trim=0 5pt 0 20pt,clip]{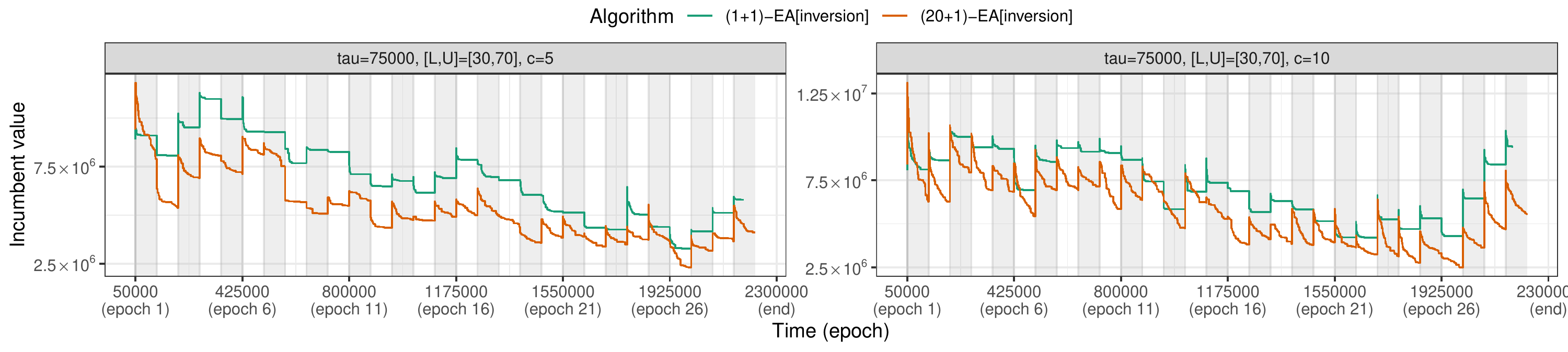}
    \caption{Representative plots of the incumbent solution trajectory of $(1+1)$-EA and $(20+1)$-EA with inversion mutation on eil101-uncorr for different dynamic setups. We show example runs for $\tau=10\,000$ (top row) and $\tau=75\,000$ (bottom row). Vertical stripes emphasise the different epochs.}
    \label{fig:trajetories_eil101}
\end{figure*}
We observe that the (1 + 1)-EA based on the inversion operator clearly outperforms the $(20+1)$-EA for all setting combinations for $\tau = 10\,000$ on eil101 and $\tau = 100\,000$ for a280. The same picture applies for the comparison between (1 + 1)-EA based on inversion and (1 + 1)-EA based on the exchange and jump operators. However, when dealing with a lower frequency of changes, i. e. $\tau = 75\,000$ or $\tau=750\,000$ respectively, we observe a phase transition where $(20+1)$-EA based on jump mostly outperforms the (1 + 1)-EA among considered mutation operators and categories of dynamic \wdtsps. The observations are statistically significant at a significance level of $\alpha = 0.05$. Pairwise Wilcoxon-Mann-Whitney tests (with Bonferroni $p$-value adjustment to account for multiple-testing issues) confirm significant superiority of $(1+1)$-EA[Inversion] with respect to mean performance in the low-frequency regime while $(20+1)$-EA dominates in the setting with large frequency.
Fig.~\ref{fig:boxplots_deviation_distribution} visualizes the performance values for all $\tau$ values (the big-picture) on instance eil101-uncorr and different dynamic setups. This figure is representative in the sense that plots for other instances and setups follow the same trends. We can clearly capture the trend observed from aggregated values in Table~\ref{tab:all_results}: population-based EAs are disadvantaged in the high frequency regions with $\tau \leq 10\,000$, but benefit from population diversity for $\tau \geq 50\,000$ where in particular $(20+1)$-EA with jump mutation consistently outperforms the $(1+1)$-EA competitors. A close look reveals that for $\tau \geq 50\,000$ the median performance of $(20+1)$-EA[jump] is below zero, i.e. in $50\%$ of the runs the algorithm beats the offline baseline.

Mean values also sometimes fall below zero (see the corresponding column in Table~\ref{tab:all_results}). 
Fig.~\ref{fig:trajetories_eil101} shows representative trajectories on eil101-uncorr where we plot the incumbent value, i.e. best so far tour length, against time for epoch lengths $\tau \in \{10\,000, 75\,000\}$. The aforementioned phase transition is obvious. A close look at the bottom row reveals how the $(1+1)$-EA converges after each dynamic change quite fast (after approximately ten to twenty thousand iterations). In contrast, the $(20+1)$-EA continues its downward trend until the onset of the next epoch.
%
\section{Conclusion}
\label{sec:sec7}

Dynamic optimization problems play a crucial role in many applications. In this paper, we have presented a dynamic setup for the recently introduced node-weighted Traveling Salesperson Problem~(\wdtsps{}) and investigated the impact of dynamic changes on the resulting optimized tours. Furthermore, we have examined evolutionary algorithms optimizing the dynamic \wdtsps{} and considered a wide range of dynamic settings in dependence of the magnitude and changes of the items. Our results show that populations are helpful for dealing with the studied dynamic problems if the parameter $\tau$ denoting the frequency of changes is sufficiently large. It should be noted that this dynamic setup is directly applicable to other packing problems such as the classical knapsack problem or the packing part of the traveling thief problem. Therefore, we expect that the dynamic setup will also be useful for studying dynamic variants of these important combinatorial optimisation problems.


\bibliographystyle{unsrt}
\bibliography{bib}

\end{document}